\documentclass[runningheads]{llncs}
\usepackage[T1]{fontenc}
\usepackage{graphicx,verbatim}
\usepackage{multirow}
\usepackage{booktabs}
\usepackage{bbding}
\usepackage{amssymb}
\usepackage{amsmath}
\begin{document}
\title{CoGE: Sim-to-Real Online Geometric Estimation for Monocular Colonoscopy}
%
\author{Liangjing Shao\inst{1,2} \and
Beilei Cui\inst{1} \and
Hongliang Ren\inst{\footnotemark[1],1,2}}
\authorrunning{L. Shao et al.}
%
\institute{Department of Electronic Engineering, The Chinese University of Hong Kong, Hong Kong SAR, China \and
Shenzhen Loop Area Institute, China \\
\email{\{leonking-shaw, beileicui\}@link.cuhk.edu.hk, hlren@ee.cuhk.edu.hk}}

\footnotetext[1]{Corresponding Author}

  
\maketitle              
\begin{abstract}
Geometric estimation including depth estimation and scene reconstruction is a crucial technique for colonoscopy which can provide surgeons with 3D spatial perception and navigation. However, geometric ground truth in colonoscopy is difficult to obtain due to narrow and enclosed space of the colon, while there is a large feature gap between simulated data and realistic data caused by artifacts and illumination. In this paper, we present CoGE, a novel framework for online monocular geometric estimation during colonoscopy. Firstly, we propose an illumination-aware supervision module based on the Retinex theory to address illumination diversity in different colonoscopy scenes. Moreover, a structure-aware perception module is proposed based on wavelet decomposition to extract common structural and local features of the colon. Both quantitative and qualitative results demonstrate that the proposed model solely trained on simulated data achieves state-of-the-art performance in geometric estimation for both simulated and realistic scenes.


\end{abstract}
\section{Introduction}
Geometric estimation from monocular endoscopic videos is crucial for 3D perception, spatial navigation and path planning in colonoscopy, due to 2D vision of endoscope and long narrow closed pathway of colon.

Recently, numerous 3D reconstruction foundation models \cite{spa,str,monst,cut,endo3r} have been proposed for cross-scene geometric estimation. However, existing methods usually need a large amount of realistic data with geometric ground truth including depth map, point clouds and camera parameters. Such datasets with geometric ground truth are difficult and costly to obtain in realistic colonoscopy due to the narrow and closed space of colon. Thus, several self-supervised learning-based methods \cite{ca,pcc} have been developed for colonoscopy geometric estimation. However, the performance of self-supervised models is limited compared with supervised models. Nevertheless, geometric ground truth can be obtained by simulation software such as VR-Caps \cite{vr,simcol}. Therefore, sim-to-real geometric estimation has become a potential approach for 3D perception in endoscopic scenes. Currently, some existing works \cite{s2r1,s2r2} have proposed sim-to-real methods for endoscopy geometric estimation. However, the proposed pipeline is evaluated after transforming the realistic data into the simulation-like data by a style translation network. Hence, the inference efficiency will be decreased due to data preprocessing, while inherent gaps between diverse realistic data and simulated data may degrade the cross-scene performance of appearance transformation models.

\begin{figure}
    \centering
    \includegraphics[width=\linewidth]{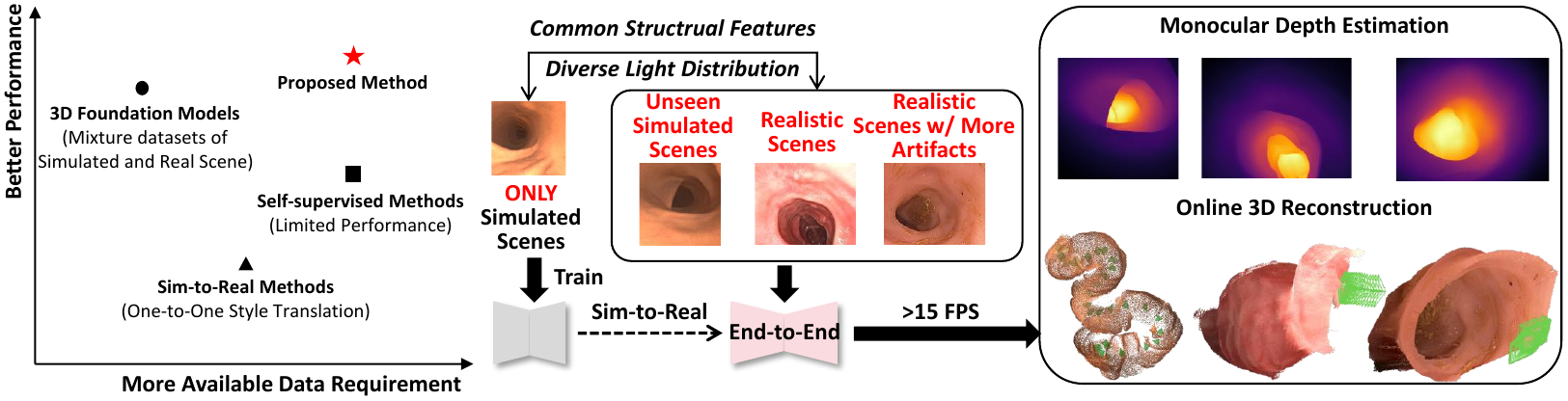}
    \caption{Given observations from realistic and unseen scenes, our model only trained on simulated data outputs depth maps and 3D point clouds.}
    \label{teaser}
\end{figure}

In this paper, we propose a foundation model-based framework, \textbf{CoGE}, for sim-to-real \textbf{g}eometric \textbf{e}stimation in \textbf{co}lonoscopy. Firstly, to extract common structure features of colon, a structure-aware perception module based on wavelet transformation is proposed to extract local and edge features in colonoscopy images. Moreover, features of colon surface are easily interfered by illumination, so a illumination-aware supervision module is proposed to control the confidence of the estimated results based on light distribution. Besides, considering long distance of colonoscopy sequence, we propose a memory forget mechanism based on attention map between feature of current frame and memory cache to pretend interference of redundant memory. 

Our main contribution can be summarized as the following: \textbf{1)} A structure-aware perception block is embedded based on wavelet transformation to extract common structural features in both simulated and realistic scenes. \textbf{2)} An illumination-aware supervision module is proposed to control the confidence of the estimated results to pretend the influence of diverse illumination on colon surface. \textbf{3)} A memory forget mechanism is proposed to filter redundant irrelevant memory for current observation against interference. \textbf{4)} The proposed model only trained on simulated data achieves state-of-the-art performance for geometric estimation on both unseen simulated scenes and realistic scenes.

\section{Method}
\begin{figure}
    \centering
    \includegraphics[width=\linewidth]{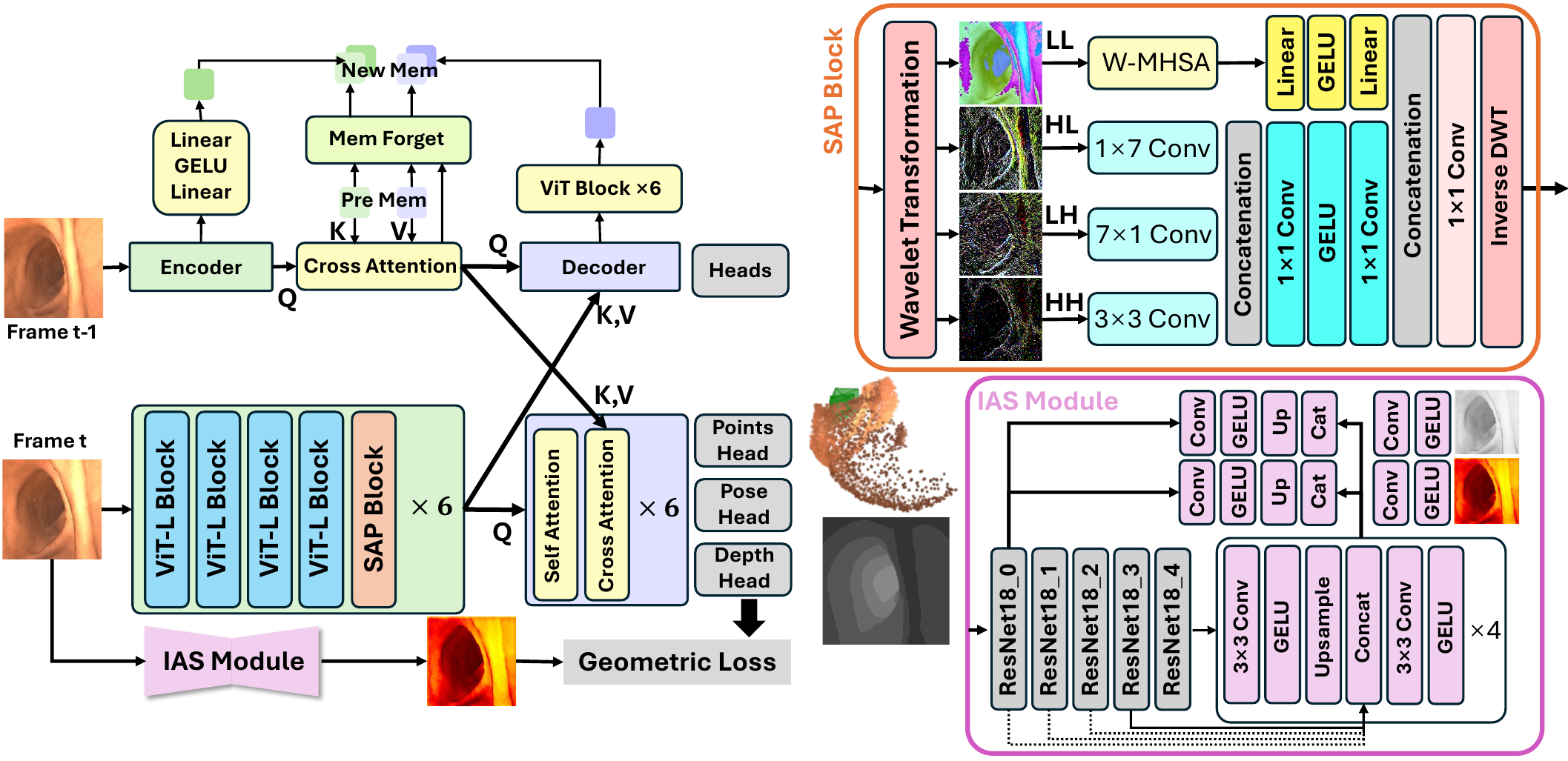}
    \caption{The pipeline of the proposed framework.}
    \label{pipeline}
\end{figure}
\subsection{Overview}
Given a pair of observations including cached previous observation $I_{t-1}$ and current observation $I_t$, the proposed model will output the corresponding point map $X_t$, camera pose $P_t$, illumination map $L_t$ and confidence map $C_t$. Meanwhile, inspired by Spann3R \cite{spa}, two spatial memory caches $M_k, M_v$ are set and they will be updated to $\hat{M_k},\hat{M_v}$ after memory reading, memory forgetting and memory updating based on the observations. The model architecture is demonstrated in Fig. \ref{pipeline}.

\subsection{Structure-aware Encoding}
Firstly, we utilize six ViT-based modules to extract scene features from the observations $f_{t-1}=\mathcal{F}_{enc}(I_{t-1}),f_{t}=\mathcal{F}_{enc}(I_{t})$. Each module consists of four ViT-L blocks followed by a structure-aware perception (SAP) block, which could extract common structural features from colonoscopy images. In SAP block, we firstly utilize discrete wavelet transformation to decompose the input graph into four subgraphs including low-low $I_{ll}$ with low-frequency main approximation of the input, low-high $I_{lh}$ with horizontal high-frequency details, high-low $I_{hl}$ with vertical high-frequency details and high-high $I_{hh}$ with diagonal high-frequency details. Convolution can extract high-frequency local details while attention is better for low-frequency information extraction. Thus, we utilize window-based multi-head attention (W-MHSA) to extract features $f_{ll}$ from $I_{ll}$, while vertical-band convolution with $7\times1$ kernel, horizontal-band convolution with $1\times7$ kernel and diagonal convolution with $3\times3$ kernel are respectively utilized to extract features $f_{hl},f_{lh},f_{hh}$ from $I_{hl}$, $I_{lh}$ and $I_{hh}$. Then, features from different wavelet compositions are fused based on Eq. \ref{sap1}-\ref{sqp3}.
\begin{equation}
    f_{h} = \mathcal{F}^{1\times1}_{conv}(\mathcal{F}_{GELU}(\mathcal{F}^{1\times1}_{conv}([f_{hl},f_{lh},f_{hh}])))
    \label{sap1}
\end{equation}

\begin{equation}
    f_{l} = \mathcal{F}_{lin}(\mathcal{F}_{GELU}(\mathcal{F}_{lin}(f_{hh})))
    \label{sap2}
\end{equation}

\begin{equation}
    f_{SAP} = \mathcal{F}_{idwt}(\mathcal{F}^{1\times1}_{conv}([f_h,f_l]))
    \label{sqp3}
\end{equation}

where $\mathcal{F}_{lin}$ denotes a linear layer, $\mathcal{F}_{idwt}$ denotes inverse discrete wavelet transformation.

\subsection{Memory-related Decoding}
\textbf{Memory Reading.} The feature from previous observation $f_{t-1}$ is utilized to retrieve the related memory information from caches by Eq. \ref{mr}.
\begin{equation}
    W_{mem} = softmax(\frac{f_{t-1}M^T_k}{\sqrt{d}})
\end{equation}

\begin{equation}
    f^{mem}_{t-1} = W_{mem}M_v+f_{t-1}
    \label{mr}
\end{equation}

\begin{figure}[!htb]
    \centering
    \includegraphics[width=0.9\linewidth]{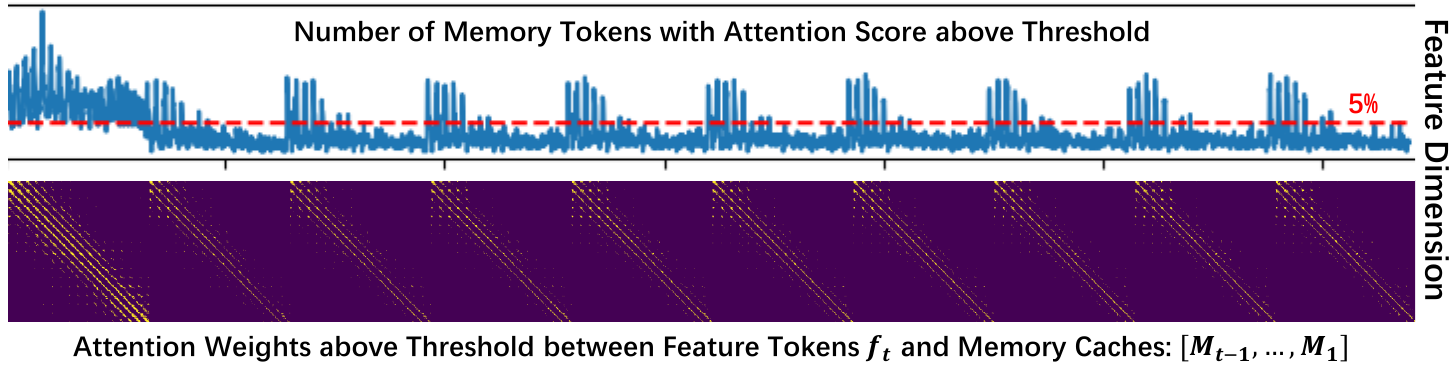}
    \caption{Attention weights distribution between feature tokens and memory caches.}
    \label{memf}
\end{figure}

\textbf{Memory Forget.} For attention maps $W_{mem}\in \mathbb{R}^{d\times s}$ between feature $f_{t-1}\in \mathbb{R}^{d\times1}$ and memory cache $M_k,M_v\in \mathbb{R}^{s\times1}$, if there is at least 5\% feature tokens $f_{t-1}(j)$ with an attention weight $W_{mem}(j,i)$ larger than 5e-4 for each memory token $M_k(i), M_v(i)$, the memory token is seemed as the relevant tokens to be retained, demonstrated as Fig. \ref{memf}.
\begin{equation}
    M_k(i),M_v(i)\rightarrow
    \begin{cases}
    retain & |\{W_{mem}(j,i)\geq\text{5e-4}\}|\geq5\%|\{W_{mem}(\cdot,i)\}|\\
    delete & \text{otherwise}
    \end{cases}
\end{equation}

\textbf{Decoding and Memory Updating.} Given memory-related feature $f^{mem}_{t-1}$ and current feature $f_t$, the output feature of current frame and previous frame can be respectively obtained by six decoding blocks, each of which consists of self-attention and cross-attention $\hat{f}_t=\mathcal{F}_{dec}(f_t,f^{mem}_{t-1}), \hat{f}_{t-1}=\mathcal{F}_{dec}(f^{mem}_{t-1}, f_t)$. The new memory of key and value are encoded from memory-related feature $\Delta M_k=\mathcal{F}^{key}_{enc}(f^{mem}_{t-1})$ and output feature of previous frame $\Delta M_v=\mathcal{F}^{val}_{enc}(\hat{f}_{t-1})$. The memory cache is updated by concatenating the new memory and filtered cached memory, $\hat{M_k},\hat{M_v}=[M_k,\Delta M_k], [M_v, \Delta M_v]$.

\textbf{Downstream Heads.} Given the output feature of the current observation $\hat{f}_t$, DPT heads $\mathcal{H}_{dpt}$ are utilized to output the corresponding point map $X_t$, camera pose $P_t$ and confidence map $C_t$, following CUT3R \cite{cut}. Furthermore, the corresponding depth map $D_t$ can be calculated based on the point map $X_t$, camera parameters $P_t$ and $K$:
\begin{equation}
    X_t,P_t,C_t=\mathcal{H}_{dpt}(\hat{f}_t),D_t=\mathcal{Z}(X_t,P_t,K)
\end{equation}

\subsection{Illumination-aware Supervision}
\begin{figure}[!htb]
    \centering
    \includegraphics[width=0.9\linewidth]{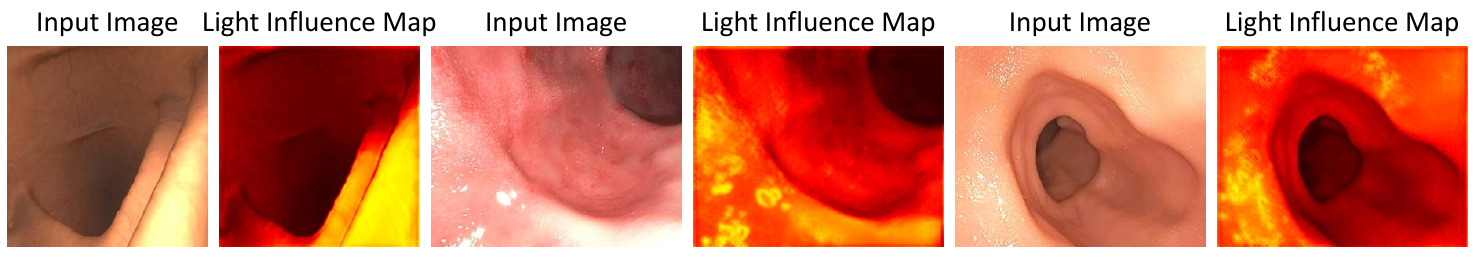}
    \caption{Input images and corresponding light influence maps.}
    \label{ias}
\end{figure}
Considering diverse and unstable illumination in colonoscopy, we propose a self-supervised illumination-aware (IAS) module to adjust confidence map for supervision. Based on retinex theory \cite{retinex}, the invariance of intrinsic component $A_t$ of image $I_t$ to illumination condition in the gradient domain ($\nabla$) is considered to generate illumination influence distribution $L_t$. Given the input image $I_t$, the IAS module outputs the corresponding intrinsic graph $A_t$ and light influence map $L_t$ shown in Fig. \ref{ias}. The IAS module is pretrained on the training split using the loss function Eq. \ref{ias} for 1 epoch with a batch size of 8. During the training of the whole network, $L_t$ from the frozen IAS module is utilized to adjust confidence map $C_t$ by Eq. \ref{adj} with a learnable weight $\alpha$.
\begin{equation}
    \mathcal{L}_{ias} = (1-L_t) \log \nabla I_t - \log \nabla A_t
    \label{ias}
\end{equation}

\begin{equation}
    \hat{C}_t=\alpha C_t+(1-\alpha) L_t
    \label{adj}
\end{equation}

Our supervision loss is mainly based on the 3D geometric loss used in CUT3R \cite{cut}, which can be formulated as Eq.\ref{loss}. 
\begin{equation}
  \mathcal{L}=  \mathcal{L}_{conf}(X_t,\hat{C}_t)+\mathcal{L}_{pose}+\mathcal{L}_{rgb}
  \label{loss}
\end{equation}

\section{Experiments and Results}
\subsection{Experiment Implementation}
We use 24 simulated sequences along with corresponding camera poses and depth maps obtained by VR-Caps \cite{vr,simcol} for training. Simulated dataset SimCol3D \cite{simcol} which includes 9 simulated scenes with 9009 frames and realistic datasets C3VD \cite{c3vd}, C3VDv2 \cite{c3vd2} including 11 sequences with over 3000 frames are utilized for evaluation. We also extract some in-house colonoscopy images for qualitative comparison. The evaluation splits mainly follow the setting in previous works \cite{pcc,ca}. The proposed model is initialized with pretrained weights from DUSt3R \cite{dust3r} and trained for 10 epochs with a batch size of 4 on a NVIDIA H100 GPU. All evaluations are implemented on a NVIDIA RTX 4090 GPU.

\begin{table}[!htb]
\centering
\caption{Quantitative results of monocular depth estimation.}
\begin{tabular}{c|c|c|ccccc|c}
\hline
 & \textbf{Methods}   & \textbf{Year}                             & \textbf{$Rel_{Abs}\downarrow$} & \textbf{$Rel_{Sq}\downarrow$} & \textbf{$RMSE\downarrow$}  & \textbf{$RMSE_{log}\downarrow$} & \textbf{$\delta\uparrow$}     & \textbf{FPS}     \\ \hline
\multirow{10}{*}{\rotatebox[origin=c]{90}{\textbf{SimCol}}}                                                   & \textbf{AF-SfM \cite{afsfm}}    & 2022 & 0.086            & 0.036           & 0.585          & 0.104             & 0.954          & 193              \\
& \textbf{DepthAnything \cite{da}}           & 2024 & 0.089            & 0.042           & 0.599          & 0.112             & 0.948          & 89               \\
& \textbf{EndoDAC \cite{endodac}}  & 2024 & 0.107            & 0.050           & 0.530          & 0.133             & 0.924          & 53               \\
& \textbf{ColonAdapter \cite{ca}}      & 2025 & 0.062            & 0.034           & 0.373          & 0.094             & 0.969          & - \\
& \textbf{MonoPCC \cite{pcc}}     & 2025 & 0.058            & 0.028           & 0.347          & 0.090             & 0.975          & - \\
& \textbf{Spann3R \cite{spa}}     & 2025 & 0.122            & 0.103           & 0.577          & 0.181             & 0.795          & 30               \\
& \textbf{MonST3R \cite{monst}}   & 2025 & 0.032            & 0.020           & 0.205          & 0.068             & 0.985          & 22               \\
& \textbf{CUT3R \cite{cut}}       & 2025 & 0.036            & 0.032           & 0.231          & 0.070             & 0.985          & 34               \\
& \textbf{STream3R \cite{str}}    & 2026 & 0.034            & 0.013           & 0.204          & 0.061             & 0.990          & 35               \\
& \textbf{Ours}                                  & 2026 & \textbf{0.027}   & \textbf{0.008}  & \textbf{0.169} & \textbf{0.049}    & \textbf{0.995} & 19               \\ \hline
\multirow{9}{*}{\rotatebox[origin=c]{90}{\textbf{C3VD}}} & \textbf{AF-SfM \cite{afsfm}}    & 2022 & 0.117            & 1.324           & 7.520          & 0.171             & 0.882          & 189              \\
& \textbf{DepthAnything \cite{da}}  & 2024         & 0.121            & 1.155           & 6.239          & 0.165             & 0.874          & 87               \\
& \textbf{EndoDAC \cite{endodac}} & 2024 & 0.103            & 0.652           & 5.306          & 0.147             & 0.890          & 50               \\
& \textbf{ColonAdapter \cite{ca}}      & 2025 & 0.139            & 0.956           & 5.592          & 0.175             & 0.832          & - \\
& \textbf{Spann3R \cite{spa}}     & 2025 & 0.090            & 0.523           & 4.152          & 0.132             & 0.891          & 26               \\
& \textbf{MonST3R \cite{monst}}   & 2025 & 0.086            & 0.541           & 4.207          & 0.134             & 0.889          & 20               \\
& \textbf{CUT3R \cite{cut}}       & 2025 & 0.087            & 0.511           & 4.144          & 0.129             & 0.896          & 29               \\
& \textbf{STream3R \cite{str}}    & 2026 & 0.085   & 0.493           & 4.105          & 0.130             & 0.900          & 30               \\
& \textbf{Ours}                    & 2026               & \textbf{0.083}   & \textbf{0.476}  & \textbf{4.061} & \textbf{0.126}    & \textbf{0.913} & 18               \\ \hline
\multirow{5}{*}{\rotatebox[origin=c]{90}{\textbf{C3VDv2}}} & \textbf{Spann3R \cite{spa}}     & 2025 & 0.108            & 2.200           & 8.582          & 0.305             & 0.802          & 20               \\
& \textbf{MonST3R \cite{monst}}  & 2025  & 0.109            & 2.014           & 8.269          & 0.260             & 0.825          & 26               \\
& \textbf{CUT3R \cite{cut}}      & 2025 & 0.132            & \textbf{1.713}  & 9.014          & 0.275             & 0.749          & 29               \\
& \textbf{STream3R \cite{str}}    & 2026 & 0.124            & 2.090           & 9.205          & 0.274             & 0.765          & 30               \\
& \textbf{Ours}                   & 2026                & \textbf{0.098}   & 2.008           & \textbf{7.772} & \textbf{0.250}    & \textbf{0.865} & 18 \\ \hline  
\end{tabular}
\label{qr}
\end{table}

\subsection{Results}
\begin{figure}[!htb]
    \centering
    \includegraphics[width=\linewidth]{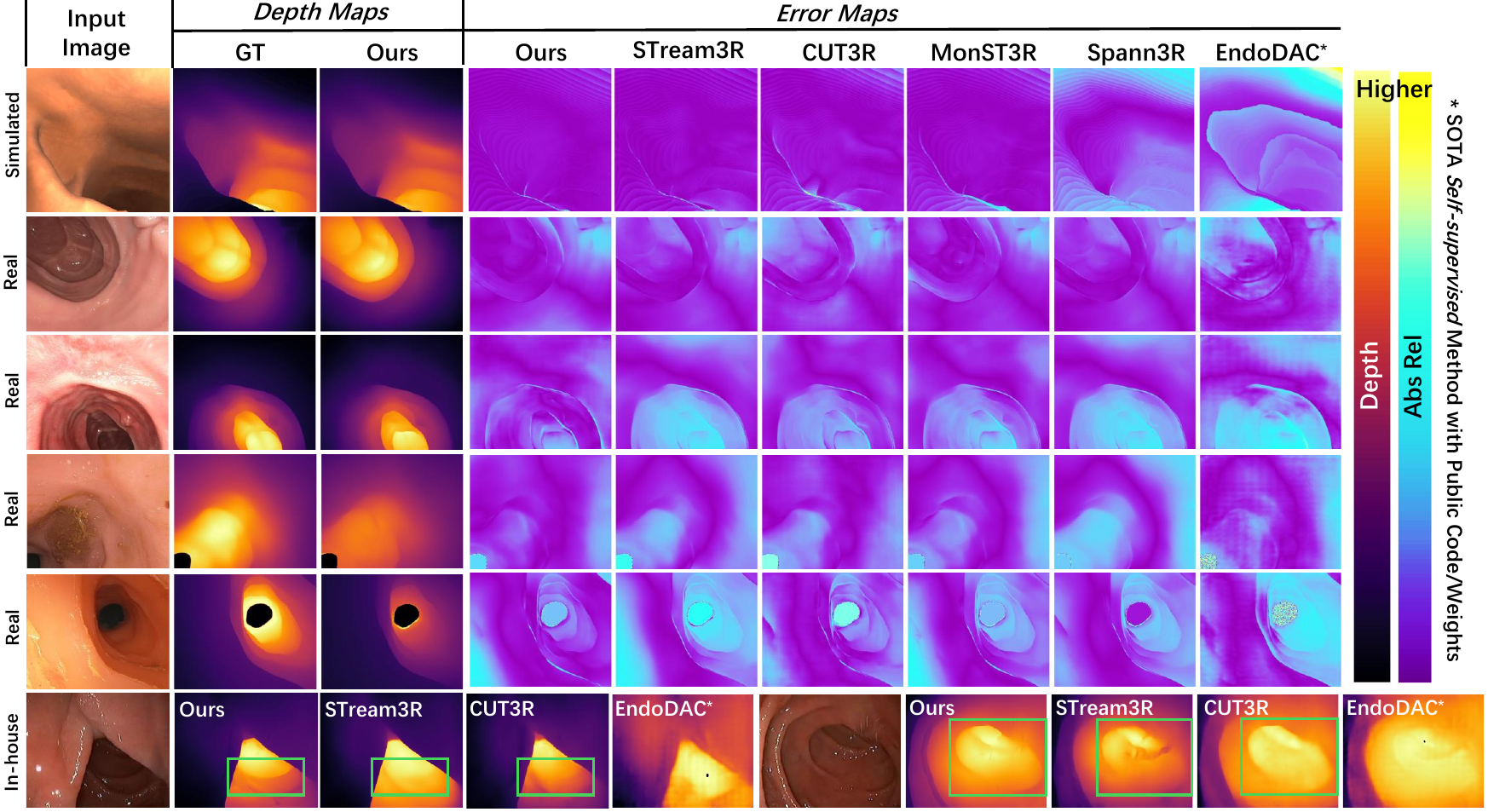}
    \caption{Qualitative results of monocular depth estimation.}
    \label{vis1}
\end{figure}

\textbf{Monocular Depth Estimation.} The proposed method is compared with SOTA 3D geometric estimation methods on both simulated data and realistic data for monocular depth estimation. Following previous works \cite{afsfm,endodac}, we use Abs Rel ($Rel_{Abs}$), Sq Rel ($Rel_{Sq}$), RMSE, RMSE Log ($RMSE_{log}$) and $\delta<1.25$ ($\delta$) as evaluation metrics. Results of MonoPCC \cite{pcc} and ColonAdapter \cite{ca} are directly from public papers, while others are trained for similar time based on the same implementation by our own. Quantitative results in Table \ref{qr} and qualitative results in Fig. \ref{vis1} demonstrate the outstanding performance of our model on monocular depth estimation, only with training on simulated data.

\begin{figure}[!htb]
    \centering
    \includegraphics[width=\linewidth]{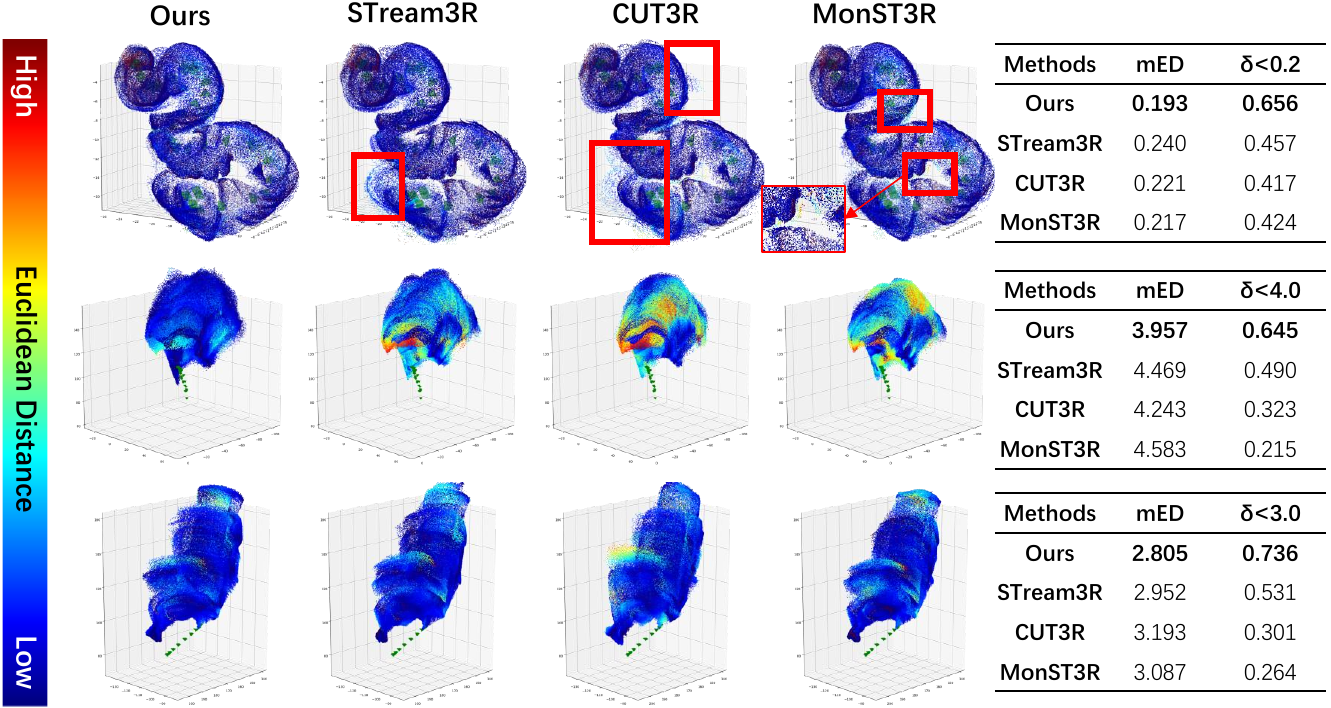}
    \caption{Qualitative Comparison of 3D reconstruction.}
    \label{vis2}
\end{figure}

\begin{figure}[!htb]
    \centering
    \includegraphics[width=0.8\linewidth]{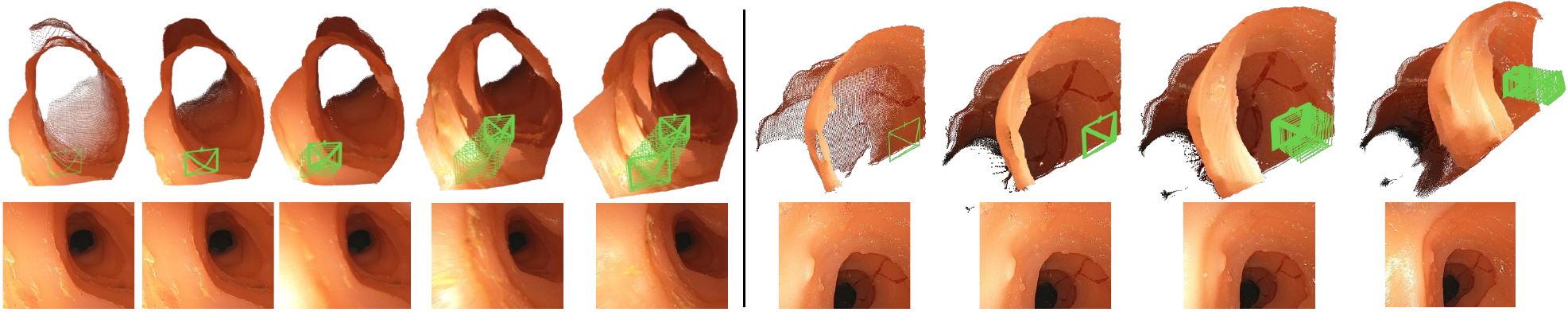}
    \caption{Qualitative Display of Our 3D reconstruction.}
    \label{vis3}
\end{figure}

\textbf{3D Reconstruction.} The proposed method is also compared with several 3D foundation models for 3D reconstruction based on monocular videos. We utilize mean Euclidean Distance (mED) to the point cloud generated from ground truth and ratio of points with ED<$N$ mm ($\delta<N$) to evaluate the reconstruction accuracy. Results in Fig.\ref{vis2} demonstrate that our method provides more accurate point cloud reconstruction compared with existing reconstruction methods. Fig. \ref{vis3} further displays some example results from our method.

\begin{figure}[!htb]
    \centering
    \includegraphics[width=0.85\linewidth]{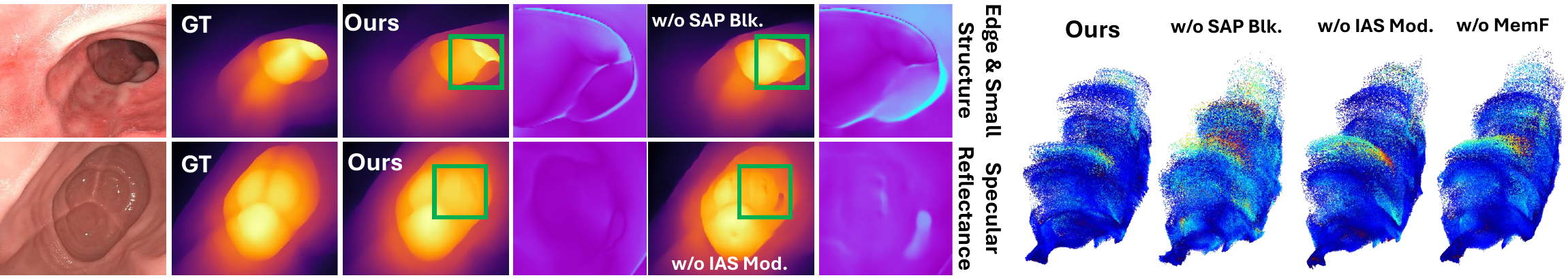}
    \caption{Qualitative results of ablation studies.}
    \label{qab}
\end{figure}

\textbf{Ablation Studies.} Our main contributions include memory forget gate (MemF), illumination-aware supervision module (IAS) and structure-aware perception Block (SAP) for challenges of sim-to-real geometric estimation in colonoscopy. Results shown in Table. \ref{ab} and Fig. \ref{qab} demonstrate the effectiveness of each main component we propose in this paper.

\begin{table}[!htb]
\fontsize{8}{8}\selectfont
\centering
\caption{Results of Ablation Studies}
\begin{tabular}{ccc|cc|cc|cc}
\hline
\multicolumn{3}{c|}{\textbf{Ablations}}                                                                                                                                                   & \multicolumn{2}{c|}{\textbf{SimCol}} & \multicolumn{2}{c|}{\textbf{C3VD}} & \multicolumn{2}{c}{\textbf{C3VDv2}} \\ \hline
\textbf{MemF} & \textbf{IAS} & \textbf{SAP} & \textbf{Abs Rel}   & \textbf{$\delta$}      & \textbf{Abs Rel}  & \textbf{$\delta$}     & \textbf{Abs Rel}  & \textbf{$\delta$}      \\ \hline
\CheckmarkBold                                                             & \CheckmarkBold                                                           & \CheckmarkBold                                                           & \textbf{0.024}     & \textbf{0.994}  & \textbf{0.085}    & \textbf{0.913} & \textbf{0.098}    & \textbf{0.865}  \\
×                                                             & \CheckmarkBold                                                           & \CheckmarkBold                                                           & 0.026              & 0.991           & 0.087             & 0.905          & 0.100             & 0.861           \\
\CheckmarkBold                                                             & ×                                                           & \CheckmarkBold                                                           & 0.030              & 0.988           & 0.087             & 0.900          & 0.104             & 0.853           \\
\CheckmarkBold                                                             & \CheckmarkBold                                                           & ×                                                           & 0.032              & 0.993           & 0.089             & 0.903          & 0.099             & 0.860           \\
\hline
\end{tabular}
\label{ab}
\end{table}

\section{Conclusion}
In this paper, we propose a geometric estimation model, which is only trained on simulated data, to perform online depth estimation and 3D reconstruction in realistic colonoscopy. We focus on illumination diversity and common structure feature extraction for sim-to-real geometric estimation. Our method outperforms state-of-the-art methods based on quantitative and qualitative experimental results on realistic data, with effects of our contributions. Despite of real-time inference speed, the inference efficiency of the proposed model is still decreased due to extra structure-aware perception blocks. Moreover, the performance of the proposed model will be degraded due to motion blur, acute dynamic change and specific lesion tissue in colonoscopy.


%
%
%
\bibliographystyle{splncs04}
\bibliography{references}
\end{document}